\documentclass{article}

\usepackage{arxiv}
\usepackage{times}
\usepackage[utf8]{inputenc}
\usepackage[T1]{fontenc}
\usepackage{amsmath,amssymb,mathtools}
\usepackage{url}
\usepackage{xcolor}
\usepackage{graphicx}
\usepackage{tikz}
\usetikzlibrary{arrows.meta}
\usepackage{booktabs}
\usepackage{colortbl}
\usepackage[expansion=false,nopatch=eqnum]{microtype}

 \usepackage[numbers,sort&compress]{natbib}
\usepackage{adjustbox}

\definecolor{linkblue}{rgb}{0.21,0.49,0.74}
\usepackage[breaklinks,colorlinks,allcolors=linkblue]{hyperref}

\makeatletter
\newcommand{\arxiv@fittabular}[2][c]{\ifdim\tabcolsep<4pt \setlength{\tabcolsep}{4pt}\fi
  \begin{adjustbox}{max width=\linewidth}\arxiv@tabular[#1]{#2}}
\newcommand{\arxiv@endfittabular}{\arxiv@endtabular\end{adjustbox}}
\let\arxiv@table\table
\renewcommand{\table}{\let\scriptsize\small
  \let\footnotesize\small
  \let\arxiv@tabular\tabular
  \let\arxiv@endtabular\endtabular
  \let\tabular\arxiv@fittabular
  \let\endtabular\arxiv@endfittabular
  \arxiv@table}
\renewenvironment{table*}{\table}{\endtable}
\makeatother

\title{AR-WAM: Efficient Streaming World-Action Modeling for Autonomous Driving}

\newcommand{\projectpage}{https://baitingluo.github.io/ar-wam/}
\newcommand{\orglogos}{\includegraphics[height=21pt]{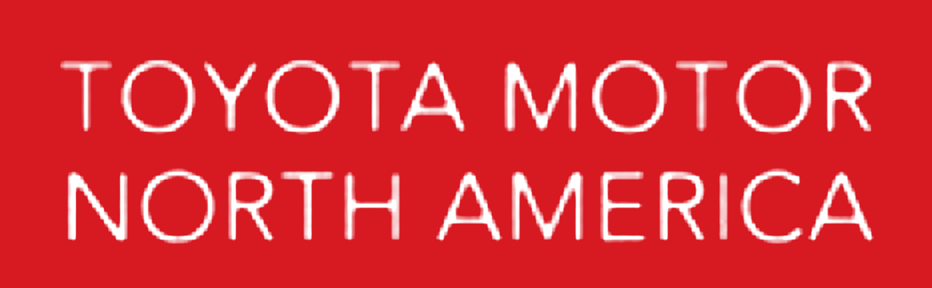}\hspace{4pt}\includegraphics[height=21pt]{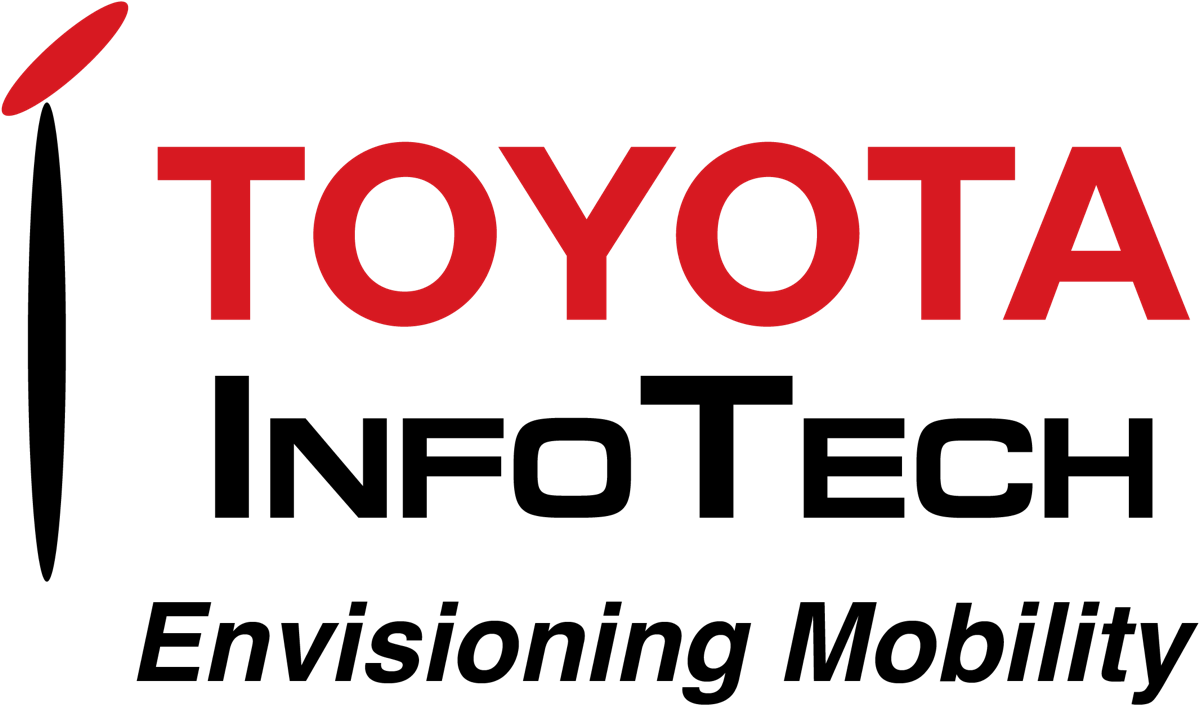}}
\makeatletter
\let\arxiv@endabstract\endabstract
\renewcommand{\endabstract}{\par\medskip\noindent
  \if\relax\detokenize\expandafter{\projectpage}\relax\else
    \textbf{Project page:} \expandafter\url\expandafter{\projectpage}\fi
  \hfill\raisebox{\dimexpr-.5\height+.35em\relax}{\orglogos}\arxiv@endabstract}
\makeatother

\author{Baiting Luo \quad Shatadal Mishra \quad Ahmadreza Moradipari \quad Nejib Ammar \\
  Toyota Motor North America R\&D - InfoTech Labs \\
  \texttt{\{baiting.luo, shatadal.mishra, ahmadreza.moradipari, nejib.ammar\}@toyota.com}}
\date{}

\hypersetup{
  pdftitle={AR-WAM: Efficient Streaming World-Action Modeling for Autonomous Driving},
  pdfauthor={Baiting Luo, Shatadal Mishra, Ahmadreza Moradipari, Nejib Ammar},
}

\begin{document}
\maketitle
\makeatletter\global\@topnum\z@\makeatother

\begin{abstract}
Autonomous driving requires anticipating how traffic will evolve while
responding quickly to new observations. Video foundation models offer the
first ability, but repeated history encoding and iterative video
generation make them costly to use for driving. We present AR-WAM, an
efficient streaming world-action model that combines autoregressive
temporal modeling with single-pass future conditioning. Because its video
backbone is causal, AR-WAM updates its visual context incrementally,
reusing past computation and conditioning future prediction on observed
motion. We further hypothesize that future features computed from recent
observations can improve generalization in dynamic traffic without
iterative video generation. AR-WAM therefore obtains these features in one
forward pass and uses them to condition trajectory generation, and
reinforcement learning then improves the policy through direct
optimization of driving rewards. With a single front camera, AR-WAM
achieves 91.5 PDMS after supervised training and 94.0 after reinforcement
learning on NAVSIM-v1, outperforming the compared strong end-to-end
world-action models. In a dynamic setting of continuous driving,
supervised AR-WAM keeps its visual history across decisions and
outperforms strong world-action baselines.
\end{abstract}
 \section{Introduction}
\label{sec:introduction}

End-to-end driving policies increasingly build on pretrained foundation
models. Vision-language-action (VLA) models transfer semantic knowledge
from image-text pretraining to trajectory prediction
\citep{li2025recogdrive,li2026drivevlaw0}. World-action models (WAMs)
instead start from video foundation models, whose pretraining on how
scenes evolve provides priors over geometry, motion and near-future
dynamics
\citep{shi2026drivewam,xia2026drivelaw,liu2026driveva,zhao2026simwam}.
These priors suit driving, because the scene changes continuously and the
right action depends on how other road users are moving: a lead vehicle
at the same distance may be braking or pulling away, and a single image
often cannot tell the two apart.

A driving policy therefore needs recent observations, which reveal
changes in relative position and speed. Current-image policies, including
OpenVLA and $\pi_0$ in robot manipulation
\citep{kim2024openvla,black2025pi0}
and current-frame driving WAMs \citep{zhao2026simwam}, omit this visual
history, although memory-based policies demonstrate its value on
temporally dependent tasks \citep{shi2026memoryvla}. Driving WAMs such as
DriveVA and DriveLaW do encode a window of recent frames
\citep{liu2026driveva,xia2026drivelaw}. With bidirectional temporal
attention, however, historical features depend on the other frames in the
window, so each new observation requires re-encoding the window
\citep{yin2025causvid}. More history then buys motion information at the
cost of computation repeated at each decision.

Future prediction adds a second source of latency. WAMs such as DriveWAM
and DriveVA iteratively predict future video before or alongside action
generation \citep{shi2026drivewam,liu2026driveva}, so each decision pays
for video denoising even when, as in DriveWAM, the observed history is
cached. To avoid this cost, Fast-WAM and SimWAM remove future video tokens
at inference while retaining video supervision during training
\citep{yuan2026fastwam,zhao2026simwam}. Yet Simple-WAM shows, in controlled
comparisons on robot manipulation, that retaining future conditioning
improves generalization, and that a single forward pass over fully noised
future tokens recovers most of the benefit \citep{zhou2026simplewam}. Motivated
by these findings, we seek a driving policy that uses both temporal
history and predictive features without repeating observation encoding or
generating future video iteratively.

\begin{figure*}[t]
\centering
\includegraphics[width=\linewidth]{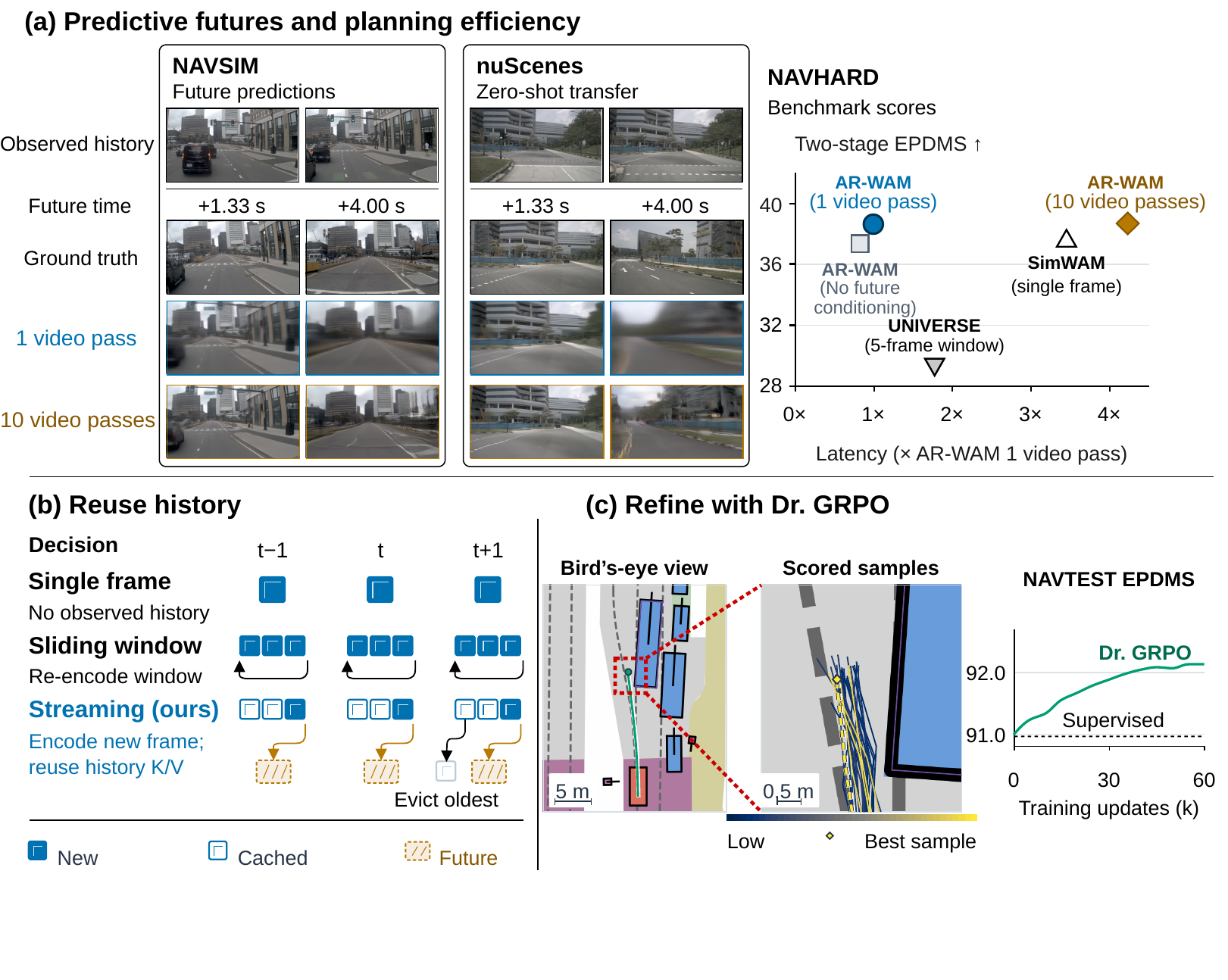}
\caption{AR-WAM plans from cached visual history and a single predictive
pass. (a) Predicted futures and NAVHARD EPDMS against latency. One video
pass scores above the variant without future conditioning (38.6 versus
37.3) and matches ten video passes (38.8) at under a quarter of their
latency, while SimWAM and UNIVERSE score lower at higher latency.
(b) AR-WAM encodes only the new frame and reuses its cached history,
whereas a sliding-window policy re-encodes its window at every decision.
(c) Reward refinement with Dr.~GRPO raises NAVTEST EPDMS to 92.2; the
scored samples show 32 plans for one scene, colored by reward.}
\label{fig:teaser}
\end{figure*}

We present AR-WAM, an efficient streaming world-action model that combines
autoregressive temporal modeling with single-pass future conditioning
(Fig.~\ref{fig:teaser}). Because causal attention makes the keys and
values (K/V) of completed observation chunks reusable when new frames
arrive, as in streaming video generation
\citep{yin2025causvid,huang2025selfforcing}, AR-WAM updates its visual
history incrementally and makes it available to future prediction.
Autoregressive pretraining also learns how future states depend on past
observations, and recent manipulation results show that it can strengthen
the benefit of visual history \citep{huang2026longwam}. We therefore ask
how much test-time future prediction contributes once the policy already
observes recent motion, and how much of it a driving decision needs.

At each decision, AR-WAM computes only what has changed. It adds the new
observation to its cache and looks ahead with one backbone pass over the
current image and fully noised future tokens. Its trajectory expert then
reads the resulting predictive features and the cached history at matching
backbone depths, so no future video is denoised. The ego state and driving command also
change at every decision, and conditioning the history on them would
require recomputing it each time \citep{huang2026longwam}. AR-WAM
therefore applies them only to the predictive pass and the expert, which
keeps the cache valid as driving intent changes. We train video and
trajectory prediction jointly, and reinforcement learning then improves
the policy through direct optimization of driving rewards
\citep{liu2025flowgrpo,zhao2026simwam}.

Built on OmniDreams, a pretrained causal video backbone
\citep{nvidia2026omnidreams}, with a single front camera, AR-WAM attains a
higher NAVSIM-v1 PDMS than strong WAMs that rely on bidirectional video
backbones, such as SimWAM, which plans from a single observation, and
UNIVERSE, which re-encodes a window of frames \citep{liu2026universe}. In
a dynamic setting of continuous driving, it also outscores both while
running 3.4 and 1.8 times faster, and visual memory accounts for most of
its lead over SimWAM. With this history available, one predictive pass
matches ten passes of joint future denoising at under a quarter of their
latency, and its benefit shows on the perturbed NAVHARD scenes. After
refinement with the EPDMS reward, AR-WAM reaches the highest NAVSIM-v2
EPDMS on both NAVTEST and NAVHARD among the compared WAMs that predict a
single trajectory without scoring candidates at test time.

Our contributions are:
\begin{itemize}
\item We introduce AR-WAM, a causal world-action model that plans from an
incrementally updated cache of observed history and a single predictive
pass. Two designs keep this cache reusable: the ego state and driving
command enter only the per-decision pass, and decision-relative attention
lets one trajectory expert read cached and predictive features at any
point of a stream.
\item We study the role of test-time future prediction once visual history
is available, and integrate it into autoregressive world-action modeling
as a single backbone pass over noised future tokens.
\item We demonstrate strong single-camera planning performance, achieving
91.5 NAVSIM-v1 PDMS after supervised training and 94.0 after reinforcement
learning. In the dynamic setting, AR-WAM scores higher than the compared
world-action models at lower latency.
\end{itemize}
 \section{Related work}
\label{sec:related}

\paragraph{Future modeling at test time.}
World-action models are trained to predict the future, but they differ in
how much of it they compute when they act. Generating future video before
or alongside the trajectory \citep{shi2026drivewam,liu2026driveva} makes
every decision pay for iterative denoising. To avoid this latency,
Fast-WAM, SimWAM and UNIVERSE keep video prediction as a training signal
and remove the future tokens at inference
\citep{yuan2026fastwam,zhao2026simwam,liu2026universe}. The action head
then reads no predictive features, however, and Simple-WAM and Faster-WAM
show in manipulation that this costs generalization
\citep{zhou2026simplewam,zhao2026fasterwam}. Other methods keep predictive
features while avoiding a full rollout at every decision
\citep{xia2026drivelaw,zhang2025epona,zhang2026rift,fan2026drivehwm,ang2026adaptivewam},
and Simple-WAM finds that a single pass over pure-noise future tokens
recovers most of the loss. Yet it and Adaptive-WAM establish this only for
policies that see the current observation, so it remains unclear how such
a pass should work beside a cache of history and how much it adds once the
policy observes recent motion. Building on the single-pass insight, AR-WAM
computes predictive features in one pass that attends to cached
observation history, and compares it with a variant without future tokens
and with full denoising.

\paragraph{Visual history and memory.}
The right action depends on motion that a single image does not show.
SimWAM and Adaptive-WAM nevertheless plan from the current frame
\citep{zhao2026simwam,ang2026adaptivewam}, and so cannot observe how other
road users are moving. DriveVA, DriveLaW and UNIVERSE observe motion
through a window of recent frames
\citep{liu2026driveva,xia2026drivelaw,liu2026universe}, but because the
window is encoded again whenever a frame arrives, their cost grows with
the history they keep. Caching the K/V of observed frames removes this
repetition. DriveWAM caches observed video and action chunks
\citep{shi2026drivewam}, yet it generates future video at every decision,
and on NAVSIM it makes a single prediction from the current frame, so its
memory plays no role there. In manipulation, Long-WAM reuses the K/V of
observed frames within one action solve and notes what would invalidate
them across decisions, such as an updated robot state or shifted temporal
positions \citep{huang2026longwam}, whereas MemoryVLA++ retrieves history
from a separate memory bank and still denoises future latents with a
separate world model \citep{shi2026memoryvlapp}. OmniDreams, the causal
backbone we build on, reports preliminary results for a policy that runs
it once per decision \citep{nvidia2026omnidreams}, but that policy reads
the backbone through a single token and is not described as reusing
computation across decisions. Motivated by these gaps, AR-WAM pairs a
cache of observed history with the rollout-free pass above, so that
neither history encoding nor future denoising is repeated at a decision.
Because conditioning visual features on the ego state or driving command,
as SimWAM does for the command, would invalidate the cache whenever they
change, AR-WAM computes history without these inputs and applies them only
in the per-decision pass. Because cached features keep the positions at
which they were computed, its trajectory expert measures positions
relative to the current decision. It therefore sees the same layout at the
start of a stream and after minutes of driving, and no cached feature has
to be modified.

\paragraph{Reward refinement.}
Flow-GRPO makes the sampler of a flow-matching model stochastic, so that
sampled outcomes can drive policy-gradient updates
\citep{liu2025flowgrpo}, and SimWAM applies it to driving with rewards
computed in the logged scene \citep{zhao2026simwam}. SimWAM restricts the
update to LoRA adapters of its action expert, whereas DSRL keeps the
policy frozen and trains an auxiliary policy over its initial noise
\citep{wagenmaker2025dsrl}. Building on these methods, AR-WAM refines the
full expert, which is affordable because all sampled plans share one
frozen visual pass. Motivated by the analysis of Dr.~GRPO
\citep{liu2025drgrpo}, it replaces per-group reward standardization with a
shared scale when training on mixed scenes.

Together, these gaps motivate the design principle of AR-WAM: the visual
computation of a decision never depends on the candidate trajectory, and
its history part never depends on the ego state or command. As a result,
history is reused across decisions, the predictive pass across action
samples, and both across reward-refinement updates, so one model addresses three challenges jointly.
 \section{Preliminaries}
\label{sec:preliminaries}

The policy observes a front-camera history $H=(I_{-N_H+1},\ldots,I_0)$
and a vector $q$ containing ego state, recent ego poses and the driving
command.
It predicts $A=(a_1,\ldots,a_T)$, where each $a_j$ is a future ego pose
(planar position and heading) in the current ego frame. Training
also provides the demonstrated trajectory $A^\star$ and logged
future video. $N_H$ denotes the number of observed images, $L$ the number
of blocks in each of the two networks of Sec.~\ref{sec:method}, and $K$
the number of action integration steps.

For normalized clean data $x$ and Gaussian noise $\epsilon$, flow matching
uses $x^\sigma=(1-\sigma)x+\sigma\epsilon$ and regresses the velocity
$\epsilon-x$ \citep{lipman2023flowmatching,liu2023rectifiedflow}.
The trajectory expert $v_\theta$, with parameters $\theta$, predicts this
velocity and integrates from noise to a clean normalized trajectory:
\begin{equation}
 A^{\sigma_{k+1}}=A^{\sigma_k}+(\sigma_{k+1}-\sigma_k)
 v_\theta(A^{\sigma_k},\sigma_k;c,q),
 \label{eq:euler}
\end{equation}
with $\sigma_1=1$ and $\sigma_{K+1}=0$. Here $c$ is the visual context,
the features that the expert reads from the video backbone
(Sec.~\ref{sec:context}). The returned
physical plan is $A=\operatorname{norm}^{-1}(A^0)$. Action flow time $\sigma$
and video flow time $\tau$ are noise coordinates.

\paragraph{The pure-noise prediction target.}
Let $Y=y_{1:F}$ be the clean latents of the $F$ future frames and
$\epsilon_v$ independent Gaussian noise. Video flow matching uses
$Y^\tau=(1-\tau)Y+\tau\epsilon_v$ with target velocity $\epsilon_v-Y$.
At $\tau=1$, the population minimizer of the squared-error video objective is
\begin{equation}
 \begin{aligned}
 u^\star&=\mathbb E[\epsilon_v-Y\mid\epsilon_v,H,q]\\
 &=\epsilon_v-\mathbb E[Y\mid H,q],\\
 \widehat Y^\star&=\epsilon_v-u^\star=\mathbb E[Y\mid H,q].
 \end{aligned}
 \label{eq:conditional-mean}
\end{equation}
The implied clean estimate therefore targets a conditional mean in latent
space. This result describes the video prediction itself. The trajectory
expert reads intermediate features instead, which the action loss also
trains, so Sec.~\ref{sec:future} tests whether these features help
planning and whether further denoising adds to them.
 \begin{figure*}[t]
\centering
\includegraphics[width=\linewidth]{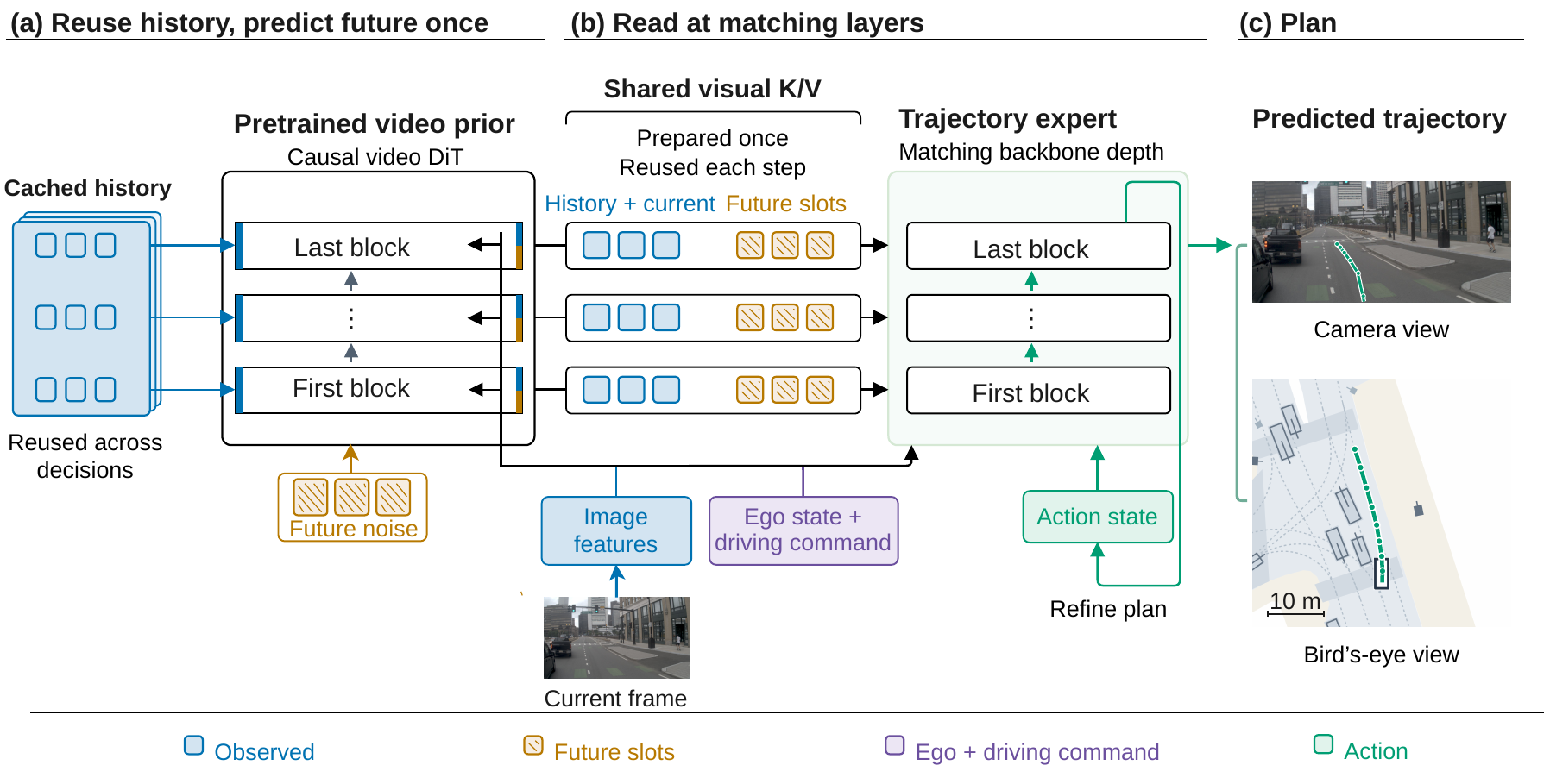}
\caption{AR-WAM architecture. (a) Each decision runs one pass of the
backbone, a causal video DiT, over the current frame and noised future slots, attending
to the cached K/V of earlier frames; the ego state and driving command
enter this pass but not the cache. (b) At matching backbone depths, the
trajectory expert reads the shared visual K/V. (c) The predicted
trajectory for a NAVTEST scene.}
\label{fig:architecture}
\end{figure*}

\section{AR-WAM}
\label{sec:method}

AR-WAM pairs a pretrained causal video diffusion transformer (DiT), the
backbone with parameters $\phi$, with the trajectory expert $v_\theta$ of
Eq.~\eqref{eq:euler}; each has $L$ blocks (Fig.~\ref{fig:architecture}).
Visual history is computed once and cached as frames arrive
(Sec.~\ref{sec:stream}), each decision adds one predictive pass
(Sec.~\ref{sec:context}), and the expert reads both at every depth
(Sec.~\ref{sec:expert}).

\subsection{Cached visual history}
\label{sec:stream}
\label{sec:streaming}

A policy that re-encodes a window of recent frames repeats most of its
visual computation at every decision. Causal video generators avoid such
repetition: because a frame never depends on later ones under causal
attention, the attention keys and values (K/V) of earlier frames can be
cached and reused \citep{yin2025causvid,huang2025selfforcing}. Building on
these designs, AR-WAM keeps its observation history in the causal form of
the backbone and bounds it with a sliding window.

Let $n$ index the frames of a stream. When frame $n$ arrives, a causal tokenizer encodes it into one latent, carrying its internal state
forward. The backbone processes the history latents $y_{\mathrm h}$ in
fixed-size chunks under chunk-causal attention: a chunk attends to
itself and at most $W$ preceding chunks, never to later tokens. The K/V of
a completed chunk therefore enter a cache
that keeps the newest $W$ chunks, and only the newest, incomplete chunk is
recomputed when a frame arrives. Positions are encoded with rotary
embeddings~\citep{su2021roformer}, under which attention depends on
positions only through the offset between query and key; each history
latent keeps its stream index as its temporal position. Each decision
then costs
$C=C_{\mathrm{inputs}}+C_{\mathrm{history}}(W)+C_{\mathrm{world}}(W)+K\,C_{\mathrm{expert}}(W)$,
with terms for encoding the new frame, updating at most one history
chunk, the predictive pass of Sec.~\ref{sec:context} and the trajectory expert's $K$
integration steps. For fixed $W$, memory and per-decision work stay
bounded however long the stream runs.

\subsection{One predictive pass per decision}
\label{sec:context}

Generating future video by iterative denoising multiplies the backbone
cost of each decision. Simple-WAM~\citep{zhou2026simplewam} finds, for
manipulation policies conditioned on the current observation, that one
forward pass over future tokens left at pure noise retains most of the
benefit of test-time future modeling. Without observation history, that
study leaves open how such a pass coexists with a cache: what it may
attend to, whether it is stored, and how per-decision inputs enter.
Building on this insight, AR-WAM introduces a world segment that is
rebuilt at each decision.

The segment $X_{\mathrm w}=\operatorname{Patch}([y_0;y^\tau_{1:F}])$ holds
patch tokens of the current latent $y_0$, encoded separately from the
history, and of $F$ future slots: the future latents $y_{1:F}$
noised to video flow time $\tau$, or one fixed Gaussian sample ($\tau=1$)
at deployment. It takes temporal positions $n-F,\ldots,n$. Its tokens
attend to all cached history K/V and follow the chunk-causal mask among
themselves. The backbone pass $\Phi_\phi$
thus yields the visual context of
Eq.~\eqref{eq:euler},
$c_\tau=\Phi_\phi(y_{\mathrm h},y_0,y^\tau_{1:F};q)$,
the K/V of history and world tokens at every block. At $\tau=1$, the video
flow objective trains the future slots toward the conditional mean of the
future given the observations (Eq.~\eqref{eq:conditional-mean}), so one
pass yields prediction-trained features without generating video.

Because the ego state and driving command $q$ change at every decision
whereas observed frames do not, conditioning the history on $q$ would
invalidate the cache at each change. AR-WAM therefore injects $q$ only
into the world segment and the expert. World tokens receive an embedding
of $q$ from a
learned ego conditioner through the backbone's conditioning
cross-attention, and history tokens always use the unconditional context
(Appendix~\ref{app:implementation}). No candidate trajectory enters the
pass either, so its K/V are shared by all action samples and integration
steps.

\subsection{Layerwise trajectory expert}
\label{sec:expert}

Pooling the backbone into a single feature vector would discard the
spatial and temporal detail that motion reasoning needs. World-action
models that pair a video backbone with a separate action network instead
let that network attend to the backbone's features inside each attention
layer \citep{yuan2026fastwam,zhao2026simwam}, reading the current frame
only. Because AR-WAM must also read a cache whose keys carry ever-growing
stream positions, we extend this layerwise attention into a
decision-relative form, in which the trajectory expert sees the same relative
positions however long the stream has run.

The expert has one block per backbone block, and its block
$\ell\in\{1,\ldots,L\}$ attends to backbone block $\ell$. From the
backbone it receives the keys and values of the history tokens,
$\mathbf K^\ell_{\mathrm h},\mathbf V^\ell_{\mathrm h}$, and of the world
tokens, $\mathbf K^\ell_{\mathrm w},\mathbf V^\ell_{\mathrm w}$, which
over all blocks form $c_\tau$. The expert adds two kinds of tokens of its
own. Action tokens embed the noisy trajectory $A^{\sigma_k}$ and give the
queries $\mathbf Q^\ell_{\mathrm a}$ and the keys and values
$\mathbf K^\ell_{\mathrm a},\mathbf V^\ell_{\mathrm a}$. Auxiliary tokens
hold higher-resolution features of the current image $I_0$ from a
separate encoder and an embedding of $q$, and give
$\mathbf K^\ell_{\mathrm x},\mathbf V^\ell_{\mathrm x}$. With
$\operatorname{Attn}(\mathbf Q,\mathbf K,\mathbf V)$ denoting attention
and $[\,\cdot\,;\cdot\,]$ concatenation over tokens, each action query
attends to all four groups:
\begin{equation}
 \begin{aligned}
 \mathbf O^\ell=\operatorname{Attn}\bigl(\mathcal R_\Delta\mathbf Q_{\mathrm a}^\ell,\;&
 [\mathbf K_{\mathrm h}^\ell;\mathbf K_{\mathrm w}^\ell;\mathcal R_\Delta\mathbf K_{\mathrm x}^\ell;\mathcal R_\Delta\mathbf K_{\mathrm a}^\ell],\\
 &[\mathbf V_{\mathrm h}^\ell;\mathbf V_{\mathrm w}^\ell;\mathbf V_{\mathrm x}^\ell;\mathbf V_{\mathrm a}^\ell]\bigr).
 \end{aligned}
 \label{eq:decision-relative}
\end{equation}
Here $\mathcal R_\Delta$ shifts the temporal position of a query or key
by $\Delta=n-F$, the stream position at which the world segment starts.
The shift reconciles two ways of counting time: backbone keys carry
positions in the stream, which keep growing, whereas the expert places
its own tokens as if every world segment started at zero. Shifting the
expert's queries and keys by $\Delta$ moves them onto the stream's axis.
Because rotary attention depends only on the offset between query and
key, the expert then sees each backbone key at its distance from the
current decision, and attention among its own tokens is unchanged. Only
the expert's side is shifted, so no cached key is modified, and the same
weights apply at the start of a stream ($\Delta=0$) and deep into a long
one. The expert integrates the $K$ Euler steps of
Eq.~\eqref{eq:euler}, reusing the decision's visual K/V at every step.

\subsection{Joint training}
\label{sec:training}

The backbone, its ego conditioner, the current-image encoder and the
trajectory expert are trained jointly with two flow-matching losses:
\begin{equation}
 \mathcal L=\mathcal L_{\mathrm{act}}+\lambda_v\mathcal L_{\mathrm{video}},
 \label{eq:video-loss}
\end{equation}
where $\mathcal L_{\mathrm{act}}$ regresses the velocity of the normalized
demonstrated trajectory $A^\star$, $\mathcal L_{\mathrm{video}}$ regresses
the velocity of the $F$ future latents through a video head used only in
training, and $\lambda_v$ weights the video term. At deployment the future
slots are pure noise, so the expert must be trained on such inputs, while
the backbone should keep video supervision at every noise level. We
therefore adopt the mixed noise schedule of
Simple-WAM~\citep{zhou2026simplewam}: all future slots share one video
flow time, set to $\tau=1$ with probability $p$ and drawn from the usual
schedule otherwise. Appendix~\ref{app:implementation} gives the losses,
schedules and weights.

\subsection{Reward refinement}
\label{sec:rl}

Imitation fits the recorded maneuver without comparing the consequences
of alternative plans. Policy gradients can make that comparison, but the
deterministic integration of a flow policy gives no tractable likelihood
for a sampled plan. Flow-GRPO~\citep{liu2025flowgrpo} injects noise into
the sampling steps of flow-matching generators, which yields exploration
and tractable path likelihoods, and SimWAM~\citep{zhao2026simwam} shows
that rewards computed in the logged scene can refine a driving expert in
this way. Building on these results, AR-WAM refines its full expert at
low cost: because the visual pass does not depend on the trajectory, each
scene needs one frozen pass, shared by all sampled plans and policy
updates. We inject Gaussian noise into the expert's $K$ integration steps
and sample $G$
plans per scene. A rule-based driving score $R$, such as the PDM
score~\citep{dauner2024navsim}, rates each plan in the scene's logged
state. Group-relative advantages compare each plan with the others from
its scene and drive a clipped policy-gradient update with a velocity
anchor to the starting expert. Such advantages are commonly divided by the
standard deviation of the rewards in their group. With rule-based driving
scores this is a poor fit, because nearly tied groups are the norm. Such
scores consist mostly of pass/fail checks, so the plans that a
well-trained policy samples in an ordinary scene all pass and differ only
slightly in score, and the division would weigh these small differences as
heavily as a collision avoided in a difficult scene.
Dr.~GRPO~\citep{liu2025drgrpo} identifies this bias toward groups whose
rewards differ little and removes the division. We do the same and train
on a mix of mined hard scenes and ordinary scenes with one shared reward
scale. At deployment, the expert integrates deterministically from a zero
trajectory.
Appendix~\ref{app:initializer-update} gives the exploration kernel, the
objective and the scene pools.
 \section{Experiments}
\label{sec:experiments}

\subsection{Evaluation protocol}
\label{sec:setup}

\paragraph{Training and evaluation.}
AR-WAM is trained on all 103,288 NAVTRAIN scenes with one front camera. We
report NAVSIM-v1 PDMS and its components, NAVSIM-v2 EPDMS and the
two-stage NAVHARD score \citep{dauner2024navsim,cao2025pseudosimulation}
on the 12,146 NAVTEST and 5,912 NAVHARD scenes. Standard NAVTEST queries
reset the streaming policy and supply four history frames. The supervised
policy plans with ten integration steps and the reward-refined policies
with four; Appendix~\ref{app:implementation} gives the training details.

\paragraph{Baselines.}
Baseline values in Table~\ref{tab:navtest} are published results, each
with its own cameras and training data. The dynamic setting of
Sec.~\ref{sec:replay} has no published results, so we run four models
that handle history and the future differently. SimWAM plans from
the current frame with the expert design closest to ours, and UNIVERSE,
the strongest peer-reviewed single-camera world-action model in
Table~\ref{tab:navtest}, re-encodes a window of frames. DriveVA
\citep{liu2026driveva} also re-encodes a window but generates future
video with the trajectory, and DriveWAM \citep{shi2026drivewam} does so
from the current frame before planning.
Appendix~\ref{app:evaluation} gives details.

\begin{table}[t]
\centering
\caption{Future modeling at inference. PDMS: NAVTEST; EPDMS: NAVHARD
two-stage; L2: nuScenes zero-shot average in metres
(Sec.~\ref{sec:nuscenes}); Passes: backbone world passes per decision;
Latency: median per decision relative to one pass, which takes 146\,ms on
an H200. Bold: best result in each column.}
\label{tab:ablation}
\footnotesize
\setlength{\tabcolsep}{3pt}
\begin{tabular}{@{}lccccc@{}}
\toprule
Future & PDMS $\uparrow$ & EPDMS $\uparrow$ & L2 $\downarrow$ & Passes & Latency \\
\midrule
No future & \textbf{91.5} & 37.3 & \textbf{0.87} & 1 & \textbf{0.8$\times$} \\
\rowcolor{gray!12} One pass (AR-WAM) & \textbf{91.5} & 38.6 & \textbf{0.87} & 1 & 1.0$\times$ \\
Clear future & \textbf{91.5} & \textbf{38.8} & 0.88 & 10 & 4.2$\times$ \\
\bottomrule
\end{tabular}
\end{table}

\begin{table*}[t]
\centering
\caption{Comparison on NAVTEST and NAVHARD. Baseline values are
publication-reported. Cam.: cameras (L: LiDAR; M: multiview). $\dagger$:
reward post-training. $\ddagger$: proposal scoring. Bold: best result in
each column, not counting the human driver and $\ddagger$ rows.}
\label{tab:navtest}
\small
\setlength{\tabcolsep}{4pt}
\begin{tabular}{@{}llccccccccc@{}}
\toprule
& & & \multicolumn{6}{c}{NAVTEST, NAVSIM-v1} & NAVTEST & NAVHARD \\
\cmidrule(lr){4-9}\cmidrule(lr){10-10}\cmidrule(l){11-11}
Method & Ref. & Cam. & NC & DAC & TTC & Comf. & EP & PDMS & EPDMS & EPDMS \\
\midrule
\multicolumn{11}{@{}l}{\emph{Reference planners without a world model}} \\
Human driver \citep{dauner2024navsim} & -- & -- & 100.0 & 100.0 & 100.0 & 99.9 & 87.5 & 94.8 & -- & -- \\
DiffusionDrive \citep{liao2025diffusiondrive} & CVPR'25 & 3+L & 98.2 & 96.2 & 94.7 & \textbf{100.0} & 82.2 & 88.1 & -- & -- \\
ReCogDrive$^\dagger$ \citep{li2025recogdrive} & ICLR'26 & 1 & 97.9 & 97.3 & 94.9 & \textbf{100.0} & 87.3 & 90.8 & -- & -- \\
DrivoR$^\ddagger$ \citep{kirby2026drivor} & CVPR'26 & 4 & 98.9 & 98.3 & 96.2 & 100.0 & 89.1 & 93.1 & -- & 48.3 \\
\midrule
\multicolumn{11}{@{}l}{\emph{World-action models, direct prediction, supervised}} \\
Epona \citep{zhang2025epona} & ICCV'25 & 1 & 97.9 & 95.1 & 93.8 & 99.9 & 80.4 & 86.2 & -- & -- \\
DriveVLA-W0 \citep{li2026drivevlaw0} & ICLR'26 & 1 & -- & -- & -- & -- & -- & -- & 86.1 & 24.4 \\
PWM \citep{zhao2025pwm} & NeurIPS'25 & 1 & 98.6 & 95.9 & 95.4 & \textbf{100.0} & 81.8 & 88.1 & -- & -- \\
DriveLaW \citep{xia2026drivelaw} & CVPR'26 & 1 & 99.0 & 97.1 & 96.7 & \textbf{100.0} & 81.3 & 89.1 & 88.6 & 30.6 \\
DriveWAM \citep{shi2026drivewam} & NeurIPS'26 & 1 & 98.3 & 98.1 & 95.2 & \textbf{100.0} & 84.3 & 90.1 & -- & -- \\
SimWAM, Wan2.2 \citep{zhao2026simwam} & arXiv & 1 & 98.6 & 98.0 & 95.8 & -- & 84.0 & 90.3 & 90.2 & 37.6 \\
DriveVA \citep{liu2026driveva} & ECCV'26 & 1 & 99.0 & 97.3 & 98.4 & \textbf{100.0} & 83.2 & 90.5 & -- & -- \\
GeoWAM \citep{lu2026geowam} & arXiv & M & -- & -- & -- & -- & -- & -- & 90.7 & 39.6 \\
SUV \citep{yuan2026suv} & arXiv & 1 & 99.1 & 97.8 & 96.7 & \textbf{100.0} & 84.6 & 90.8 & 91.0 & 36.9 \\
UNIVERSE \citep{liu2026universe} & NeurIPS'26 & 1 & 99.1 & 97.6 & \textbf{98.5} & \textbf{100.0} & 83.6 & 91.0 & -- & -- \\
WA-JEPA \citep{wang2026wajepa} & arXiv & 4 & \textbf{99.5} & 98.3 & 97.7 & \textbf{100.0} & 85.0 & 91.8 & 91.7 & -- \\
\midrule
\multicolumn{11}{@{}l}{\emph{World-action models, post-trained or proposal scoring}} \\
SimWAM, Wan2.2$^\dagger$ & arXiv & 1 & 98.5 & 99.1 & 96.1 & 99.8 & 86.5 & 91.9 & -- & -- \\
SimWAM, Cosmos2.5$^\dagger$ & arXiv & 1 & 98.8 & 99.2 & 96.4 & -- & 87.0 & 92.4 & -- & -- \\
DA-WAM$^\ddagger$ \citep{zhong2026dawam} & arXiv & 1 & 99.1 & 98.9 & 96.8 & 99.8 & 90.0 & 93.7 & 89.2 & -- \\
\midrule
\rowcolor{gray!12} AR-WAM &  & 1 & 99.4 & 98.2 & 98.0 & \textbf{100.0} & 84.5 & 91.5 & 90.9 & 38.6 \\
\rowcolor{gray!12} AR-WAM + RL, PDMS$^\dagger$ &  & 1 & 99.0 & \textbf{99.3} & 96.1 & \textbf{100.0} & \textbf{90.7} & \textbf{94.0} & -- & -- \\
\rowcolor{gray!12} AR-WAM + RL, EPDMS$^\dagger$ &  & 1 & 99.3 & \textbf{99.3} & 97.1 & \textbf{100.0} & 86.6 & 92.7 & \textbf{92.2} & \textbf{42.3} \\
\bottomrule
\end{tabular}
\end{table*}

\begin{table*}[t]
\centering
\caption{Dynamic setting: the 47,220 non-waiting frames of 138 NAVTEST
logs, run in order at 2\,Hz with a persistent history cache. Columns are
one-stage NAVSIM-v2 sub-scores. Hist.: camera frames used per decision.
Latency: median per decision at batch size one on an H200. Current frame:
AR-WAM retrained without visual history. Bold: best result in each
column.}
\label{tab:replay}
\small
\setlength{\tabcolsep}{4.5pt}
\begin{tabular}{@{}lcccccccccccc@{}}
\toprule
Method & Hist. & Latency (ms) & NC & DAC & DDC & TLC & EP & TTC & LK & HC & EC & EPDMS \\
\midrule
SimWAM \citep{zhao2026simwam} & 1 & 499.1 & 98.0 & 99.1 & 99.4 & 99.5 & 88.0 & 98.2 & 98.1 & \textbf{98.7} & 87.2 & 90.4 \\
UNIVERSE \citep{liu2026universe} & 5 & 256.9 & 98.5 & 99.0 & \textbf{99.5} & \textbf{99.8} & 87.8 & 98.7 & 97.9 & 98.6 & 64.5 & 88.3 \\
DriveVA \citep{liu2026driveva} & 5 & 258.8 & 98.5 & 99.0 & \textbf{99.5} & \textbf{99.8} & 87.7 & 98.6 & 98.2 & 98.5 & 57.4 & 87.4 \\
DriveWAM \citep{shi2026drivewam} & 1 & 895.7 & 98.0 & 99.1 & 99.4 & 99.4 & 88.3 & 98.1 & 98.3 & \textbf{98.7} & \textbf{91.8} & 90.9 \\
\midrule
AR-WAM, current frame & 1 & \textbf{106.9} & 98.2 & 99.1 & \textbf{99.5} & 99.7 & 88.0 & 98.3 & 98.5 & \textbf{98.7} & 84.8 & 90.7 \\
\rowcolor{gray!12} AR-WAM & 5 & 146.3 & \textbf{98.7} & 99.1 & \textbf{99.5} & 99.6 & 88.0 & 98.8 & \textbf{98.8} & \textbf{98.7} & 88.0 & 91.5 \\
\rowcolor{gray!12} AR-WAM + RL, EPDMS & 5 & 132.8 & \textbf{98.7} & \textbf{99.6} & \textbf{99.5} & 99.5 & \textbf{88.6} & \textbf{98.9} & 98.0 & \textbf{98.7} & 89.9 & \textbf{92.2} \\
\bottomrule
\end{tabular}
\end{table*}

\subsection{Does future modeling improve decisions?}
\label{sec:future}

Table~\ref{tab:ablation} compares the single predictive pass with two
alternatives. In the no-future variant, the trajectory expert reads only
the observed history and the current latent, although the backbone is
still trained to predict future video, as in current-frame WAMs that drop
future tokens at inference \citep{yuan2026fastwam,zhao2026simwam}. This
variant repeats AR-WAM's final training stage with the same
initialization, data order, noise draws, losses and schedule, and differs
only in what the expert reads. In the clear-future variant, AR-WAM's own
weights denoise the future video jointly with the action, following
Simple-WAM's joint inference \citep{zhou2026simplewam}, so the last action
steps read an almost clean predicted future at the cost of one backbone
pass per action step.

Test-time future prediction improves decisions at a small cost in
latency. One predictive pass raises EPDMS on the perturbed
two-stage scenes of NAVHARD by 1.3 over the no-future variant, with
unchanged NAVTEST and nuScenes accuracy, while latency grows only from
$0.8\times$ to $1.0\times$. Denoising a clear future adds a further 0.2
EPDMS but takes more than four times as long, so one pass recovers most
of its benefit at under a quarter of the latency.

\subsection{Evaluation in a dynamic setting}
\label{sec:replay}

Each standard NAVTEST query rebuilds the policy's history from four
frames, so it cannot measure the effect of history carried across
decisions.
We therefore also evaluate a dynamic setting in which each policy runs
through 138 full NAVTEST source logs in chronological order (69,711
frames at 2\,Hz) and keeps its history across frames.
Observations come from the logs and do not depend on the predicted
actions. Plans are scored with one-stage EPDMS, including extended comfort
(EC) between consecutive plans.

\paragraph{Evaluated frames.}
NAVSIM builds NAVTRAIN and NAVTEST without scenes it considers trivial,
such as standing still or driving straight at near-constant speed
\citep{dauner2024navsim}, but full logs keep them. In 32\% of the frames
in our logs, the logged vehicle moves less than 1\,m over the next 4\,s,
compared with 1.9\% of NAVTEST frames. Policies trained on NAVTRAIN therefore
rarely see sustained waiting, while UNIVERSE and DriveVA train on full
logs. To compare policies on the benchmark's scene distribution, we
exclude these waiting frames. The criterion uses only the
logged vehicle, so all policies are scored on the same frames.

On the 47,220 non-waiting frames valid for all compared policies
(Table~\ref{tab:replay}), AR-WAM outscores all four baselines and runs
faster. Its lead over SimWAM, which replans from the current frame, comes
mainly from fewer at-fault collisions and time-to-collision violations,
and visual memory accounts for most of it: the current-frame variant,
which repeats AR-WAM's final training stage without visual history, falls
between the two, closer to SimWAM. This memory costs about 40\,ms per
decision. The window models UNIVERSE and DriveVA have much lower extended
comfort, meaning less consistent consecutive plans. DriveVA, which
generates future video at every decision, scores below UNIVERSE.
DriveWAM has the highest extended comfort but trails AR-WAM by 0.6 EPDMS
at six times AR-WAM's latency. Refinement with the EPDMS reward raises both EPDMS and extended
comfort further.

\subsection{Benchmark comparison}
\label{sec:benchmarks}

Table~\ref{tab:navtest} compares AR-WAM with world-action models and
reference planners, and Table~\ref{tab:navtest-full} in the appendix
lists all compared methods. Direct methods output one trajectory per
decision; proposal-scoring methods select one of up to 64 candidates with
a scorer trained on PDM sub-scores. Among the compared supervised direct methods,
AR-WAM has the highest NAVSIM-v1 PDMS with a single camera, and only the
four-camera WA-JEPA scores higher. Its NAVSIM-v2 EPDMS is within 0.1 of
the best supervised single-camera result, and on NAVHARD only GeoWAM,
pretrained on 832 additional hours of driving video, scores higher among
direct methods. The supervised policy is cautious: its no-collision rate
is close to the best, and ego progress is its weakest term. Reward
refinement shifts this balance toward progress. With the PDMS reward,
AR-WAM reaches 94.0 PDMS, above every direct method and above the
proposal-scoring DA-WAM and DrivoR. With the EPDMS reward, it has the
best NAVSIM-v2 scores among direct methods on both NAVTEST and NAVHARD,
although DrivoR, a planner without a world model that scores its own
proposals from four cameras, stays ahead on NAVHARD.

\subsection{Zero-shot transfer}
\label{sec:nuscenes}

To test transfer beyond NAVSIM, we apply AR-WAM to nuScenes
\citep{caesar2020nuscenes} without fine-tuning, following DriveVA's
evaluation protocol \citep{liu2026driveva} on 4,519 samples from 150
validation scenes. SimWAM, UNIVERSE, DriveVA and DriveWAM are evaluated on
the same samples, together with a constant-velocity reference, since ego
state alone can explain much of open-loop performance
\citep{li2024egostatus}. Table~\ref{tab:nuscenes} also splits the samples
with the waiting criterion of Sec.~\ref{sec:replay}. NAVSIM's training
protocol removes most scenes in which the vehicle stands still, so the
non-waiting samples correspond to the kind of scenes AR-WAM is trained on.
On these samples, AR-WAM and DriveWAM transfer best, with the lowest L2
error and about half the collision rate of UNIVERSE and DriveVA. Over all
samples, including the 14\% of waiting samples that its training largely
excludes, AR-WAM stays close to UNIVERSE (0.87 versus 0.83\,m average
L2), matches DriveVA and has the lowest one-second error. DriveWAM has the
lowest collision rate but higher L2 error than AR-WAM at every horizon.
On the waiting samples, UNIVERSE and DriveVA, which train on full logs,
stay near the log, while AR-WAM outperforms SimWAM and DriveWAM, the other
NAVSIM-trained policies.

\begin{table}[t]
\centering
\caption{NAVSIM-to-nuScenes transfer. Lower is better. L2 is in metres;
Col.\ is the horizon-averaged collision percentage. The Adaptive-WAM row
is publication-reported; the others are our evaluations on
4,519 samples (Appendix~\ref{app:evaluation}). Of the 4,437 samples with
all horizons, 635 are waiting: the logged vehicle moves less than 1\,m
within 3\,s. Bold: best result in each column.}
\label{tab:nuscenes}
\scriptsize
\setlength{\tabcolsep}{2.4pt}
\begin{tabular}{@{}lrrrrrrrrr@{}}
\toprule
& \multicolumn{5}{c}{All samples} & \multicolumn{2}{c}{Non-waiting} & \multicolumn{2}{c}{Waiting} \\
\cmidrule(lr){2-6}\cmidrule(lr){7-8}\cmidrule(l){9-10}
& \multicolumn{4}{c}{L2} & Col. & L2 & Col. & L2 & Col. \\
\cmidrule(lr){2-5}
Method & 1\,s & 2\,s & 3\,s & Avg. & Avg. & Avg. & Avg. & Avg. & Avg. \\
\midrule
Constant velocity & 0.33 & 1.13 & 2.34 & 1.27 & 0.47 & 1.46 & 0.54 & \textbf{0.15} & 0.06 \\
SimWAM \citep{zhao2026simwam} & 0.29 & 0.83 & 1.77 & 0.97 & 0.05 & 0.94 & 0.05 & 1.12 & 0.05 \\
UNIVERSE \citep{liu2026universe} & 0.30 & 0.75 & \textbf{1.44} & \textbf{0.83} & 0.07 & 0.91 & 0.08 & 0.37 & \textbf{0.01} \\
DriveVA \citep{liu2026driveva} & 0.34 & 0.79 & 1.48 & 0.87 & 0.10 & 0.97 & 0.08 & 0.30 & 0.20 \\
DriveWAM \citep{shi2026drivewam} & 0.27 & 0.77 & 1.68 & 0.91 & \textbf{0.03} & \textbf{0.88} & \textbf{0.03} & 1.08 & \textbf{0.01} \\
Adaptive-WAM \citep{ang2026adaptivewam} & 0.35 & \textbf{0.71} & 1.58 & 0.88 & 0.08 & -- & -- & -- & -- \\
\rowcolor{gray!12} AR-WAM & \textbf{0.24} & 0.73 & 1.63 & 0.87 & 0.05 & \textbf{0.88} & 0.04 & 0.82 & 0.14 \\
\bottomrule
\end{tabular}
\end{table}

\subsection{Reward refinement}
\label{sec:rl-results}

Reward refinement on a mix of mined hard scenes and ordinary scenes
raises NAVTEST PDMS from 91.6 to 94.0 at four integration steps
(Table~\ref{tab:navtest}). The gain comes from ego progress and
drivable-area compliance, while time-to-collision compliance drops.
Refining with the NAVSIM-v2 EPDMS reward instead raises EPDMS from 90.9
to 92.2 on NAVTEST and from 38.6 to 42.3 on NAVHARD.

\subsection{Runtime and history reuse}
\label{sec:latency}

UNIVERSE re-encodes its five-frame window at every decision, whereas
AR-WAM encodes only the newest frame and reuses its cache. With the same
five frames, AR-WAM is 1.8 times faster than UNIVERSE and DriveVA in
median latency (Table~\ref{tab:replay}), although it also computes
predictive features. It is 3.4 times faster than SimWAM, which plans from
the current frame with ten flow steps, and 6.1 times faster than
DriveWAM. Four action steps instead of ten lower AR-WAM's latency further
at unchanged NAVTEST PDMS. Table~\ref{tab:stream-latency} in the appendix
adds tail latency and peak memory.
 \section{Conclusion}
\label{sec:conclusion}

AR-WAM pairs a cache of observed history with one predictive pass and
reads both through decision-relative attention. Because its visual
computation depends on neither the candidate trajectory nor, for history,
the ego state or command, it is reused across decisions, action samples
and reward-refinement updates. With history available, one predictive
pass matches iterative future denoising at under a quarter of its latency.
AR-WAM reaches 94.0 NAVSIM-v1 PDMS after reward refinement and leads
other strong world action baselines in the dynamic setting at lower latency.

\clearpage
\appendix
\section{Implementation details}
\label{app:implementation}

\paragraph{Architecture and inputs.}
Front-camera images are resized to $352\times640$. The frozen causal Wan
tokenizer \citep{wan2025wan} emits one latent of shape $16\times44\times80$
per observed frame, using its retained state. The current image and logged
continuation are encoded separately for the world segment, which contains
one current and three future latents ($F=3$). Spatial $2\times2$ patching
gives 880 tokens per history latent and 3,520 tokens in the world segment.
The backbone's native chunks hold two latents. History positions follow
the stream index, and world positions are relative to the newest
observation. The current-blind mask blocks access from the current latent
to its first future chunk-mate.

The video backbone has 28 blocks of width 2048, with 16 attention heads of
dimension 128. It is initialized from the public map-conditioned
OmniDreams weights \citep{nvidia2026omnidreams}. The trajectory expert has 28
blocks of width 1536 and feed-forward width 6144, using the same attention
head space. Compatible attention and feed-forward weights are initialized
by resizing backbone weights. Residual gates and the output layer start at
zero. The image branch encodes an independently tokenized $384\times672$
current image into 252 tokens of width 256.

Each backbone block cross-attends to a text context. Let $e_0$ be the
backbone's projection of a null-text row and $P_\phi$ the ego conditioner.
World tokens receive the two-row context
\begin{equation}
 E_q=[e_0;\ e_0+P_\phi(q)],
 \label{eq:ego-context}
\end{equation}
and history tokens the null context $[e_0;e_0]$. $P_\phi$ projects the
24-dimensional $q$ through a 1024-wide SiLU MLP to a 1024-dimensional offset,
with a zero-initialized final layer. It belongs to the backbone optimizer
group and is included in the exponential moving average (EMA). Using two
cross-attention rows, rather than appending the ego row to the backbone's
hundreds of identical null-text rows, keeps the ego input from being
diluted; with equal initial rows, the pretrained null-only response is
preserved exactly. As in SimWAM \citep{zhao2026simwam}, the narrower
expert projects its queries into the backbone's per-head attention space
and projects the result back to its own width; timestep-modulated residual
blocks predict the action velocity.

\paragraph{Supervised initialization.}
An initial supervised stage uses four-frame clips, a causal hold adapter
with 13 RGB slots, local four-latent rotary grids for both segments, and
ego context throughout the backbone. Its final EMA initializes streaming
training. This stage uses AdamW with $\beta=(0.9,0.95)$,
$\epsilon=10^{-8}$ and weight decay 0.01. Both parameter groups peak at
$7\times10^{-5}$ after 2,000 warm-up updates and decay by a cosine schedule
to $10^{-6}$. Global batch size is 32 on four NVIDIA H200 GPUs with
micro-batch eight.
Training runs for 25 NAVTRAIN passes, or 80,694 updates. Gradients are
clipped at norm 1.0 and activation checkpointing is enabled. EMA uses decay
0.9999 with an initial warm-up. Both flow schedules use shift $s=5$; the video
weight is $\lambda_v=3$, the endpoint probability $p=\tfrac12$ and the
endpoint video weight one. Inference uses $K=10$ flow steps with fixed
action and future-noise seeds. Plans contain $T=8$ poses over four
seconds. Coordinates are normalized using lower bounds
$(-1.57,-19.68,-1.67)$ and ranges $(66.74,42,3.53)$ for $(x,y,\psi)$;
yaw is wrapped after denormalization.

\paragraph{Losses and scheduler weights.}
For $x=\operatorname{norm}(A^\star)$, independent action noise
$\epsilon_a$ and video noise $\epsilon_v$, the two objectives in
Eq.~\eqref{eq:video-loss} are
\begin{equation}
 \begin{aligned}
 \mathcal L_{\mathrm{act}}&=
 \mathbb E\bigl[w(\sigma)\operatorname{MSE}
 (v_\theta(x^\sigma,\sigma;c_\tau,q),\epsilon_a-x)\bigr],\\
 \mathcal L_{\mathrm{video}}&=
 \mathbb E\Bigl[\tfrac{\omega(\tau)}{F}\textstyle\sum_{f=1}^{F}
 \operatorname{MSE}(u_{\phi,f},\epsilon_{v,f}-y_f)\Bigr],
 \end{aligned}
\end{equation}
where $u_{\phi,f}$ is the video head's velocity for future latent $f$.
Action flow time follows the shifted schedule
$\mathcal S_s(u)=su/(1+(s-1)u)$ with $u\sim\mathcal U(0,1)$; video flow time
is $1$ with probability $p$ and $\mathcal S_s(u_v)$ otherwise, shared by all
future latents while observed latents stay clean. Both noise regimes
contribute to both losses. Video loss updates the backbone and its ego
conditioner, while action loss also updates the current-image encoder and
the expert; the tokenizer stays frozen. In the final training stage, older
history is computed without gradients but stays in the attention context;
gradients pass through the newest history chunks and the world segment.

Let $r(t)=\exp[-2(t-\tfrac12)^2]-\exp[-\tfrac12]$ and $N=1000$. The
continuous scheduler normalizes this weight over its shifted reference grid:
\begin{equation}
 Z=\frac1N\sum_{i=1}^{N}r(\mathcal S_5(i/N)),\qquad
 w(t)=\frac{r(t)}{Z+10^{-10}}.
\end{equation}
For sampled video flow times, $\omega(\tau)=w(\tau)$. Pure-noise mixture
samples instead use $\omega(1)=1$. This is the training weight in the
released SimWAM continuous-flow scheduler, with Simple-WAM's endpoint
override \citep{zhao2026simwam,zhou2026simplewam}.

\subsection{Streaming training}
\label{app:streaming}

Streaming training starts from the initial supervised EMA at update 80,694.
The backbone, ego conditioner, current-image branch and trajectory expert are
fine-tuned jointly for 10,000 updates, or 320,000 scene presentations.
The tokenizer remains frozen. AdamW uses the same betas, epsilon and
weight decay as the initialization stage, with a peak learning rate of
$2\times10^{-5}$ after 500 warm-up updates and cosine decay to $10^{-6}$.
Global batch size is 32 on four GPUs. EMA decay is 0.9995 after its initial
warm-up. Gradient clipping, activation checkpointing and both flow-matching
objectives are unchanged. Results use the final EMA, not a selected
intermediate checkpoint.

The history cache covers all 103,288 training scenes and stores 246,165
available camera frames across 9,339 runs in 1,192 logs. A run ends at a
missing image or a timestamp gap greater than two seconds. Within a run,
the first image is encoded alone. Each later latent encodes three copies
of the preceding image followed by the new image, carrying the tokenizer's
causal state between calls. Training uses local positions within each
sample, while deployment retains stream positions. The action rotation in
Eq.~\eqref{eq:decision-relative} makes the action interface invariant to
this common temporal offset.

\subsection{Reward-based policy refinement}
\label{app:initializer-update}

\paragraph{Data and visual context.}
Refinement draws from 85,190 NAVTRAIN scenes, and 12,660 scenes from
disjoint recordings are held out for development. The training pool
combines 29,254 hard scenes, mined with an earlier supervised model, and
14,627 ordinary scenes drawn once at random from the remaining training
scenes, so hard scenes make up two thirds of the pool. Rewards are
official non-reactive NAVSIM-v1 PDM scores, computed for each candidate
against the fixed PDM-Closed reference \citep{dauner2024navsim}. A second
policy uses the NAVSIM-v2 EPDMS reward, which adds lane keeping,
traffic-light compliance and comfort between consecutive plans. Both
policies start from the supervised policy.

Each decision resets the policy and prefills four observed frames with
$W=2$. A fixed future-noise realization supplies the predictive features.
The tokenizer, video backbone, backbone ego conditioner and current-image
branch remain frozen. Each scene supplies 32 stochastic paths and one
deterministic monitoring plan; the latter is excluded from advantages and
policy losses. The full trajectory expert is trainable.

\paragraph{Short sampler and SDE exploration.}
The trajectory expert uses the reference scheduler's four-step grid,
$1=\sigma_1>\cdots>\sigma_{K+1}=0$, with $x_1=0$ and
$\Delta_k=\sigma_{k+1}-\sigma_k<0$. Let
$v_k=v_\theta(x_k,\sigma_k;c,q)$. During training, the transition is
\begin{equation}
 \begin{aligned}
 g_k&=\eta\sqrt{\sigma_k(1-\sigma_k)+\varepsilon_g},\\
 b_k&=v_k+\frac{g_k^2}{2\sigma_k}
	 \bigl(x_k+(1-\sigma_k)v_k\bigr),\\
 m_{\theta,k}&=x_k+\Delta_k b_k,\qquad
 a_k=g_k\sqrt{-\Delta_k},\\
 x_{k+1}&=m_{\theta,k}+a_k\xi_k,\quad \xi_k\sim\mathcal N(0,I_{24}).
 \end{aligned}
 \label{eq:initializer-sde}
\end{equation}
The drift follows the score-corrected flow-SDE construction used by
Flow-GRPO \citep{liu2025flowgrpo}. We use $\eta=0.018$ and
$\varepsilon_g=0.25$, giving positive variance at every transition,
including $\sigma_1=1$. The conditional density is
$p_{\theta,k}=\mathcal N(m_{\theta,k},a_k^2I_{24})$. Noise amplitudes remain
fixed during rollout reuse. Training uses this discrete stochastic policy;
deployment sets $\eta=0$, recovering $x_{k+1}=x_k+\Delta_kv_k$.

\paragraph{Scene-centered rewards.}
For the $G=32$ noisy rewards, define
\begin{equation}
 \widehat{\mathcal A}_i=\frac{R_i-\bar R}{d_R},
 \qquad \bar R=G^{-1}\sum_{j=1}^{G}R_j.
 \label{eq:initializer-advantage}
\end{equation}
We use the fixed scale $d_R=s_R=0.1$, following Dr.~GRPO's removal of
group-specific standardization \citep{liu2025drgrpo}. A common denominator
preserves the size of reward differences across scenes, so nearly tied
plans receive smaller advantages than plans separated by large reward
differences. Groups with constant rewards have zero advantages.

\paragraph{Clipped updates and reference anchor.}
For a stored path $\zeta_i=(x_{i,2},\ldots,x_{i,K+1})$ with $x_{i,1}=0$,
the conditional log density is
\begin{equation}
 \ell_i(\theta)=\sum_{k=1}^{K}\log p_{\theta,k}
 (x_{i,k+1}\mid x_{i,k},c,q),
 \label{eq:joint-policy}
\end{equation}
which sums $3T=24$ coordinates at each of the $K=4$ transitions. We temper
its change before clipping:
\begin{equation}
 \rho_i(\theta)=\exp\!\left[
 \frac{\ell_i(\theta)-\ell_i^{\mathrm{old}}}{24K}\right],
 \label{eq:policy-ratio}
\end{equation}
where $\ell_i^{\mathrm{old}}$ is the behavior log density recorded at
sampling. This is the geometric-mean density ratio over 96 transition
coordinates. The loss is
\begin{equation}
 \begin{aligned}
 \mathcal L_{\mathrm{RL}}(\theta)
 &=-\mathbb E_i\bigl[\min\{\rho_i\widehat{\mathcal A}_i,\\
 &\quad\operatorname{clip}(\rho_i,1-\delta,1+\delta)
 \widehat{\mathcal A}_i\}\bigr]\\
 &\quad+\lambda_{\mathrm{ref}}\mathcal L_{\mathrm{ref}}(\theta).
 \end{aligned}
 \label{eq:rl-loss}
\end{equation}
The reference penalty anchors velocities on the stored states:
\begin{equation}
 \mathcal L_{\mathrm{ref}}(\theta)=
 \mathbb E_{i,k}\operatorname{MSE}\!\left(
 v_\theta(x_{i,k},\sigma_k;c,q),
 v_{\theta_0}(x_{i,k},\sigma_k;c,q)\right).
 \label{eq:expert-anchor}
\end{equation}
The teacher $\theta_0$ is the expert from which refinement starts. It, the
sampled states, old log densities, rewards and frozen visual features are
detached. Gradients reach $\theta$ through
transition densities and the reference penalty. The expert's auxiliary
projections are recomputed during each optimizer pass.

\paragraph{Optimizer and budgets.}
We use AdamW with learning rate $10^{-6}$, betas $(0.9,0.95)$, zero weight
decay and 100 warm-up updates. The clipping threshold is $\delta=0.02$,
the anchor coefficient is $\lambda_{\mathrm{ref}}=100$, and the gradient
norm is clipped at 1.0. Four workers, one per GPU, collect two scenes each. Each rollout
batch supports four optimizer updates, with the same states, old
likelihoods and rewards reused. We report the PDMS-reward policy after
40,000 updates and the EPDMS-reward policy after 45,000.

\section{Evaluation details}
\label{app:evaluation}

\begin{table*}[t]
\centering
\caption{All compared methods on NAVTEST and NAVHARD, with the columns and
notes of Table~\ref{tab:navtest}. $\circ$: concurrent work. $\dagger$:
reward or preference post-training;
the DriveReferee row is its distilled single-pass policy. $\ddagger$:
proposal scoring. Bold: best result in each column, not counting the
human driver and proposal-scoring rows.}
\label{tab:navtest-full}
\footnotesize
\setlength{\tabcolsep}{4pt}
\begin{tabular}{@{}llccccccccc@{}}
\toprule
& & & \multicolumn{6}{c}{NAVTEST, NAVSIM-v1} & NAVTEST & NAVHARD \\
\cmidrule(lr){4-9}\cmidrule(lr){10-10}\cmidrule(l){11-11}
Method & Ref. & Cam. & NC & DAC & TTC & Comf. & EP & PDMS & EPDMS & EPDMS \\
\midrule
\multicolumn{11}{@{}l}{\emph{Reference planners without a world model}} \\
Human driver \citep{dauner2024navsim} & -- & -- & 100.0 & 100.0 & 100.0 & 99.9 & 87.5 & 94.8 & -- & -- \\
DiffusionDrive \citep{liao2025diffusiondrive} & CVPR'25 & 3+L & 98.2 & 96.2 & 94.7 & \textbf{100.0} & 82.2 & 88.1 & -- & -- \\
ReCogDrive$^\dagger$ \citep{li2025recogdrive} & ICLR'26 & 1 & 97.9 & 97.3 & 94.9 & \textbf{100.0} & 87.3 & 90.8 & -- & -- \\
DrivoR$^\ddagger$ \citep{kirby2026drivor} & CVPR'26 & 4 & 98.9 & 98.3 & 96.2 & 100.0 & 89.1 & 93.1 & -- & 48.3 \\
\midrule
\multicolumn{11}{@{}l}{\emph{World-action models, direct prediction, supervised}} \\
Epona \citep{zhang2025epona} & ICCV'25 & 1 & 97.9 & 95.1 & 93.8 & 99.9 & 80.4 & 86.2 & -- & -- \\
DriveVLA-W0 \citep{li2026drivevlaw0} & ICLR'26 & 1 & -- & -- & -- & -- & -- & -- & 86.1 & 24.4 \\
PWM \citep{zhao2025pwm} & NeurIPS'25 & 1 & 98.6 & 95.9 & 95.4 & \textbf{100.0} & 81.8 & 88.1 & -- & -- \\
DriveLaW \citep{xia2026drivelaw} & CVPR'26 & 1 & 99.0 & 97.1 & 96.7 & \textbf{100.0} & 81.3 & 89.1 & 88.6 & 30.6 \\
DriveWAM \citep{shi2026drivewam} & NeurIPS'26 & 1 & 98.3 & 98.1 & 95.2 & \textbf{100.0} & 84.3 & 90.1 & -- & -- \\
SV-WAM \citep{wang2026svwam} & arXiv & 6 & 98.6 & 98.6 & 95.1 & \textbf{100.0} & 83.9 & 90.2 & 91.0 & 36.1 \\
SimWAM, Wan2.2 \citep{zhao2026simwam} & arXiv & 1 & 98.6 & 98.0 & 95.8 & -- & 84.0 & 90.3 & 90.2 & 37.6 \\
DriveVA \citep{liu2026driveva} & ECCV'26 & 1 & 99.0 & 97.3 & 98.4 & \textbf{100.0} & 83.2 & 90.5 & -- & -- \\
GeoWAM \citep{lu2026geowam} & arXiv & M & -- & -- & -- & -- & -- & -- & 90.7 & 39.6 \\
SUV \citep{yuan2026suv} & arXiv & 1 & 99.1 & 97.8 & 96.7 & \textbf{100.0} & 84.6 & 90.8 & 91.0 & 36.9 \\
4D-WAM \citep{fu20264dwam} & arXiv & 1 & 99.1 & 97.9 & 96.6 & \textbf{100.0} & 84.6 & 90.9 & 90.6 & 35.9 \\
UNIVERSE \citep{liu2026universe} & NeurIPS'26 & 1 & 99.1 & 97.6 & \textbf{98.5} & \textbf{100.0} & 83.6 & 91.0 & -- & -- \\
ReDrive$^\circ$ \citep{zhu2026redrive} & arXiv & 1 & 99.1 & 97.9 & 97.2 & \textbf{100.0} & 84.3 & 91.0 & 90.8 & 34.4 \\
PhysWAM$^\circ$ \citep{parikh2026physwam} & arXiv & 3 & 99.0 & 97.5 & \textbf{98.5} & 99.8 & 88.7 & 91.4 & 90.3 & 38.1 \\
WA-JEPA \citep{wang2026wajepa} & arXiv & 4 & \textbf{99.5} & 98.3 & 97.7 & \textbf{100.0} & 85.0 & 91.8 & 91.7 & -- \\
\midrule
\multicolumn{11}{@{}l}{\emph{World-action models, direct prediction, post-trained}} \\
VITAL$^{\dagger\circ}$ \citep{liang2026vital} & arXiv & 1 & 98.7 & 98.5 & 95.8 & \textbf{100.0} & 84.8 & 90.8 & 91.1 & 38.6 \\
SimVTA$^{\dagger\circ}$ \citep{jing2026simvta} & arXiv & 1 & 98.0 & 97.2 & 97.0 & \textbf{100.0} & 86.7 & 91.3 & 88.2 & -- \\
AffordDrive3D$^{\dagger\circ}$ \citep{cai2026afforddrive3d} & arXiv & 3 & 98.6 & 99.0 & 96.3 & \textbf{100.0} & 84.7 & 91.3 & 89.9 & -- \\
DriveUniWAM$^{\dagger\circ}$ \citep{xu2026driveunitok} & arXiv & 1+L & 98.7 & 98.0 & 95.6 & 99.8 & 87.4 & 91.5 & -- & -- \\
SimWAM, Wan2.2$^\dagger$ & arXiv & 1 & 98.5 & 99.1 & 96.1 & 99.8 & 86.5 & 91.9 & -- & -- \\
DriveReferee$^{\dagger\circ}$ \citep{yu2026drivereferee} & arXiv & 1 & -- & -- & -- & -- & -- & 92.0 & 91.6 & -- \\
DriveWLA$^{\dagger\circ}$ \citep{wang2026drivewla} & arXiv & 1 & 99.1 & 98.8 & 97.0 & \textbf{100.0} & 86.4 & 92.3 & 90.1 & -- \\
SimWAM, Cosmos2.5$^\dagger$ & arXiv & 1 & 98.8 & 99.2 & 96.4 & -- & 87.0 & 92.4 & -- & -- \\
\midrule
\multicolumn{11}{@{}l}{\emph{World-action models, proposal scoring}} \\
Adaptive-WAM$^\dagger$ \citep{ang2026adaptivewam} & arXiv & 1 & 98.6 & 97.9 & 95.6 & 100.0 & 85.1 & 90.8 & 89.9 & -- \\
MORE$^\circ$ \citep{gu2026more} & arXiv & 3 & 98.8 & 98.4 & 95.6 & 100.0 & 90.7 & 93.2 & 92.5 & -- \\
DA-WAM \citep{zhong2026dawam} & arXiv & 1 & 99.1 & 98.9 & 96.8 & 99.8 & 90.0 & 93.7 & 89.2 & -- \\
PlanWAM$^\circ$ \citep{xu2026planwam} & arXiv & 4 & 98.9 & 98.8 & 96.3 & 100.0 & 91.0 & 93.8 & 90.9 & -- \\
MM-Future$^\circ$ \citep{liu2026mmfuture} & arXiv & 4 & 98.7 & 99.0 & 95.8 & 100.0 & 91.6 & 94.0 & 91.5 & -- \\
\midrule
\rowcolor{gray!12} AR-WAM &  & 1 & 99.4 & 98.2 & 98.0 & \textbf{100.0} & 84.5 & 91.5 & 90.9 & 38.6 \\
\rowcolor{gray!12} AR-WAM + RL, PDMS$^\dagger$ &  & 1 & 99.0 & \textbf{99.3} & 96.1 & \textbf{100.0} & \textbf{90.7} & \textbf{94.0} & -- & -- \\
\rowcolor{gray!12} AR-WAM + RL, EPDMS$^\dagger$ &  & 1 & 99.3 & \textbf{99.3} & 97.1 & \textbf{100.0} & 86.6 & 92.7 & \textbf{92.2} & \textbf{42.3} \\
\bottomrule
\end{tabular}
\end{table*}

\paragraph{Baselines and reported policies.}
Table~\ref{tab:navtest-full} lists every method we compare against.
Table~\ref{tab:navtest} includes the peer-reviewed world-action models
and, from earlier preprints, the methods closest to AR-WAM in design or
with the best aggregate scores. Concurrent work ($\circ$) appears only in
Table~\ref{tab:navtest-full}. All baseline values are published results,
and the reference planners' values come from their own papers.
DriveLaW's NAVSIM-v2 values and DriveVLA-W0's
NAVHARD value are taken from SimWAM \citep{zhao2026simwam}, whose
supervised Wan2.2 and RL Cosmos2.5 rows come from its backbone ablation,
which does not report comfort. Each baseline keeps its published training
setup: SimWAM uses 85,109 training samples, MM-Future and
SV-WAM also train on NAVVAL, and the DriveVA row is its NAVSIM-only model
(DriveVA reaches 90.9 PDMS when CARLA data are added)
\citep{liu2026mmfuture,wang2026svwam,liu2026driveva}. In the Cam.\
column, M marks multiview input without a stated camera count. AR-WAM's
two RL rows are separate policies, refined with the PDMS or the EPDMS
reward, and both plan with four integration steps. Our NAVSIM-v2 scores apply
the official human-reference filter of the NAVSIM devkit, and DrivoR's
scores are those of its NAVTRAIN-only model, with NAVHARD under the same
filter \citep{kirby2026drivor}. In the dynamic
setting and on nuScenes, SimWAM, UNIVERSE, DriveVA and DriveWAM run their
released
supervised checkpoints with their native inputs. The single released
DriveVA checkpoint is labeled 90.9 PDMS and scores 90.1 on NAVTEST with
our scorer, and the released DriveWAM checkpoint scores 90.1, matching its
reported value. UNIVERSE does not report NAVHARD; its score in
Fig.~\ref{fig:teaser} is our evaluation of the released checkpoint.

\paragraph{nuScenes.}
Our evaluation under DriveVA's protocol uses the validation keyframes that
have four past and at least six future keyframes. Ground-truth poses use the
rear-axle ego frame at front-camera timestamps. The command follows the final
future heading with an $8^\circ$ threshold, as in the reference evaluation.
L2 error is interpolated at 1, 2 and 3\,s. An oriented ego box is rasterized
against annotations on a 0.1\,m grid, with its centre 1.49\,m ahead of the
rear axle and heading from displacement (zero below 1\,m). Logged-colliding
steps are excluded from predicted-collision counts according to that
protocol, and collision rates average keyframes up to each horizon.
Velocity and acceleration come from CAN signals. Scoring uses DriveVA's
released code. Its final sample filter is not released, so our 4,519
samples may differ from the published sample sets. On them,
the released SimWAM and UNIVERSE checkpoints reproduce their reported
average L2 of 0.96 and 0.83\,m within 0.01\,m, the released DriveVA
checkpoint gives 0.87\,m against a reported 0.84\,m, and the released
DriveWAM checkpoint gives 0.91\,m against the 0.96\,m that SimWAM reports
for it \citep{zhao2026simwam}. SimWAM's average collision
rate is within 0.01 points of its reported value, whereas UNIVERSE's is
0.07\% against a reported 0.03\%.

\paragraph{Dynamic setting.}
Each decision uses the logged history at its time. Other vehicles
may react during scoring, but predicted sensor observations are not fed into
the next policy input. Anchor-frame scores can differ from standalone
evaluation because extended comfort pairs consecutive plans of the stream.
After waiting frames are excluded, each remaining frame keeps the score
computed in the continuous stream, including extended comfort relative to
the preceding plan.

\section{Runtime validation}
\label{app:runtime}

The accelerated implementation preserves the current-blind mask,
ego-conditioned world cross-attention and width-1536 trajectory expert. A
static per-layer ring holds $2W$ history latents. Separate regions hold
world, auxiliary, action and open-latent K/V. FlashAttention reads the
valid ranges without concatenating the cache at every action step. Split
attention outputs are merged with their log-sum-exp normalizers. Timestep
modulations are precomputed, and CUDA graphs capture completed-chunk,
world and action updates. Inductor compiles the block computations.
History cross-attention to identical null rows has a constant output,
while the ego-conditioned world cross-attention remains query-dependent.
The last backbone block supplies only the K/V consumed by the expert.
Cached decisions agree with recomputation of the same causal prefix,
including the context that formed retained states.

Both runtimes use BF16 computation; the accelerated path also stores BF16
weight copies. On a 1,024-scene NAVTEST subset with cold four-frame inputs
and batch size four, the accelerated runtime scores 91.704 PDMS against
91.705 for the reference runtime with the same checkpoint. Mean per-scene
trajectory separation is 1.28\,cm, with a 90th percentile of 2.47\,cm and
a maximum per-scene mean of 5.16\,cm.

\begin{table}[t]
\centering
\caption{Per-decision latency, peak GPU memory and NAVTEST PDMS at batch
size one on an H200. Hist.: camera frames used per decision, which
AR-WAM reads from its cache. Future: future-video computation at
inference. Baseline PDMS values are publication-reported. Bold: best
result in each column. $^\dagger$Three cameras and LiDAR.
$^\ddagger$Publication-reported mean on an A100.}
\label{tab:stream-latency}
\footnotesize
\setlength{\tabcolsep}{3pt}
\begin{tabular}{@{}lllrrrr@{}}
\toprule
& Hist. & & \multicolumn{2}{c}{Latency (ms)} & Mem. & \\
\cmidrule(lr){4-5}
Method & (fr.) & Future & Med. & P90 & (GB) & PDMS \\
\midrule
DiffusionDrive$^\dagger$ & 1 & none & \textbf{40.7} & \textbf{52.5} & \textbf{0.3} & 88.1 \\
SimWAM & 1 & dropped & 499.1 & 933.9 & 13.8 & 90.3 \\
Adaptive-WAM$^\ddagger$ & 1 & early exit & 170 & -- & -- & 90.8 \\
UNIVERSE & 5 & dropped & 256.9 & 327.2 & 24.4 & 91.0 \\
UNIVERSE & 5 & 3 steps & 341.8 & 350.9 & 24.4 & -- \\
DriveVA & 5 & 3 steps & 258.8 & 316.3 & 24.4 & 90.9 \\
DriveWAM & 1 & rollout & 895.7 & 905.4 & 26.1 & 90.1 \\
\midrule
\rowcolor{gray!12} AR-WAM, $K{=}10$ & 5 & 1 pass & 146.3 & 154.8 & 22.5 & 91.5 \\
\rowcolor{gray!12} AR-WAM, $K{=}4$ & 5 & 1 pass & 132.8 & 137.3 & 22.5 & \textbf{91.6} \\
\bottomrule
\end{tabular}
\end{table}

For Table~\ref{tab:stream-latency}, each method runs the first 40 frames
of three log segments after model loading and compilation, giving 115 to
117 timed decisions per configuration after warm-up, on an otherwise idle
GPU. Memory is the peak PyTorch allocation, including weights. Adaptive-WAM
has no released code, so its reported A100 mean is quoted. Timing starts from
decoded RGB frames in host memory and ends with the trajectory on the CPU,
so it includes resizing, encoding, the ego vector and action sampling.
AR-WAM issues the newest frame's three encodes concurrently on separate
CUDA streams; its plans agree bitwise with sequential encoding on all 120
frames for every cached history and step count. Baselines use their released NAVSIM settings. UNIVERSE takes
five history frames at $480\times832$ and three flow steps, either
trajectory-only or jointly with future video; the joint row excludes
decoding the predicted frames to pixels, which its released code adds and
which raises the median to 732\,ms. DriveVA takes the same five frames and
denoises future video latents together with the trajectory in three flow
steps without decoding them to pixels. DriveWAM takes the current front
frame at $256\times448$, rolls out one future video latent in three steps
and then the trajectory in ten. SimWAM takes the current front frame
and ten flow steps. DiffusionDrive takes the current left, front and right
cameras and two denoising steps. Our NAVTEST sensor data contain no LiDAR,
so DiffusionDrive receives a synthetic 100,000-point sweep, which enters
only through a fixed-size BEV histogram.

AR-WAM + RL uses the same network as AR-WAM and plans with four steps, so
its latency is that of the $K{=}4$ row in Table~\ref{tab:stream-latency}.
 
\end{document}